\documentclass[10pt,twocolumn,letterpaper]{article}

\usepackage{iccv}
\usepackage{times}
\usepackage{epsfig}
\usepackage{graphicx}
\usepackage{amsmath}
\usepackage{amssymb}
\usepackage{multirow}
\usepackage{caption}
\usepackage{subcaption}

\graphicspath {{./figures/}, {./figures/nyu_results/}}

\iccvfinalcopy 

\ificcvfinal\fi
\begin{document}

\title{From Big to Small: Multi-Scale Local Planar Guidance \\ for Monocular Depth Estimation}

\author{Jin Han Lee, Myung-Kyu Han, Dong Wook Ko and Il Hong Suh\\
	Department of Electronics and Computer Engineering, Hanyang University\\
	{\tt\small \{jinhanlee, mkhan91, pumpblack, ihsuh\}@hanyang.ac.kr}
}

\maketitle

\begin{abstract}	
Estimating accurate depth from a single image is challenging because it is an ill-posed problem as infinitely many 3D scenes can be projected to the same 2D scene.
However, recent works based on deep convolutional neural networks show great progress with plausible results.
The convolutional neural networks are generally composed of two parts: an encoder for dense feature extraction and a decoder for predicting the desired depth.
In the encoder-decoder schemes, repeated strided convolution and spatial pooling layers lower the spatial resolution of transitional outputs, and several techniques such as skip connections or multi-layer deconvolutional networks are adopted to recover the original resolution for effective dense prediction.

In this paper, for more effective guidance of densely encoded features to the desired depth prediction, we propose a network architecture that utilizes novel local planar guidance layers located at multiple stages in the decoding phase.
We show that the proposed method outperforms the state-of-the-art works with significant margin evaluating on challenging benchmarks.
We also provide results from an ablation study to validate the effectiveness of the proposed method.	
\end{abstract}

\section{Introduction}
Depth estimation from 2D images has been studied in computer vision for a long time and is nowadays applied to robotics, autonomous driving cars, scene understanding, and 3D reconstruction.
Those applications usually utilize, to perform depth estimation, multiple instances of the same scene such as stereo image pairs \cite{scharstein2002taxonomy}, multiple frames from moving camera \cite{ranftl2016dense} or static captures under different lighting conditions \cite{abrams2012heliometric,basri2007photometric}.
As depth estimation from multiple observations achieves impressive progress, it naturally leads to depth estimation with a single image since it demands less cost and constraint.

However, estimating accurate depth from a single image is challenging, even for a human, because it is an ill-posed problem as infinitely many 3D scenes can project to the same 2D scene.
To understand geometric configuration from a single image, humans consider use not only local cues, such as texture appearance in various lighting and occlusion conditions, perspective, or relative scales to the known objects, but also global context, such as entire shape or layout of the scene \cite{howard2012perceiving}.

After the first learning-based monocular depth estimation work from Saxena et al. \cite{saxena2006learning} was introduced, considerable improvements have been made along with rapid advances in deep learning \cite{eigen2014depth,eigen2015predicting,li2017two,liu2016learning,wang2015towards,roy2016monocular,kim2016unified,laina2016deeper}.
While most of the state-of-the-art works apply models based on deep convolutional neural networks (DCNNs) in \textit{supervised} fashion, some works proposed \textit{semi-} \cite{kuznietsov2017semi} or \textit{self-supervised} learning methods which do not entirely rely on the ground truth depth data.

\begin{figure}
	\centering
	\smallskip
	\begin{subfigure}{.3\linewidth}
		\centering
		\includegraphics[width=\linewidth]{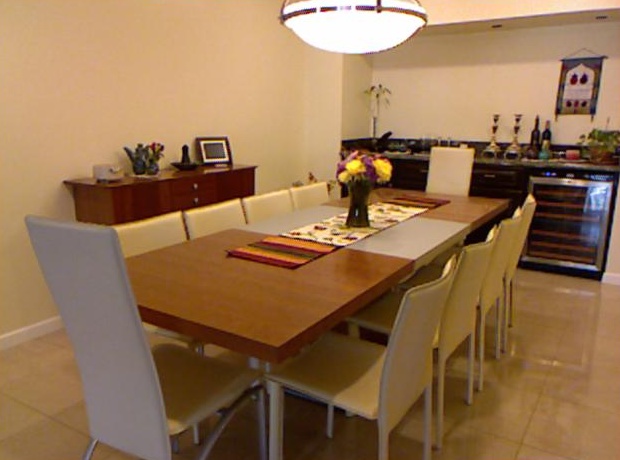}
	\end{subfigure}
	\begin{subfigure}{.3\linewidth}
		\centering
		\includegraphics[width=\linewidth]{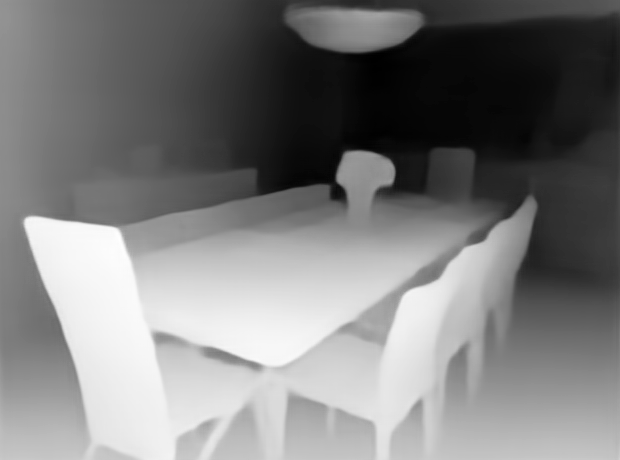}
	\end{subfigure}
	\begin{subfigure}{.3\linewidth}
		\centering
		\includegraphics[width=\linewidth]{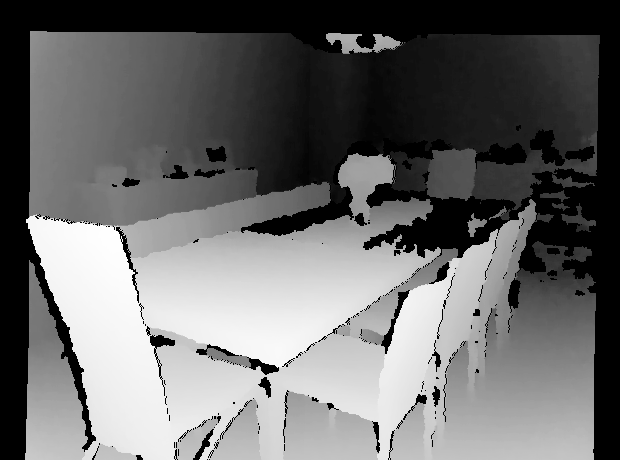}
	\end{subfigure}	
	
	\vspace{0.1cm}
	
	\begin{subfigure}{.3\linewidth}
		\centering
		\includegraphics[width=\linewidth]{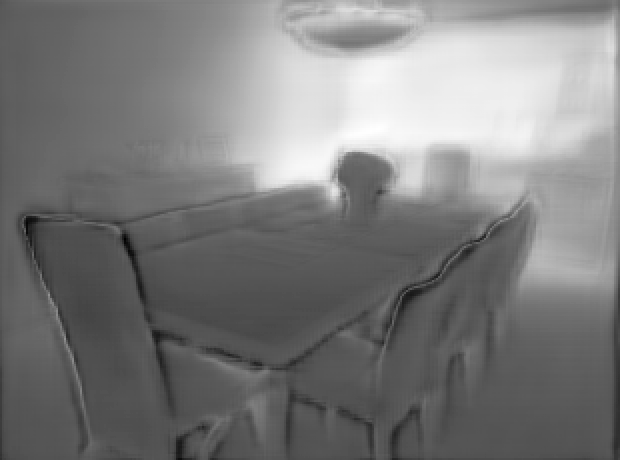}
	\end{subfigure}
	\begin{subfigure}{.3\linewidth}
		\centering
		\includegraphics[width=\linewidth]{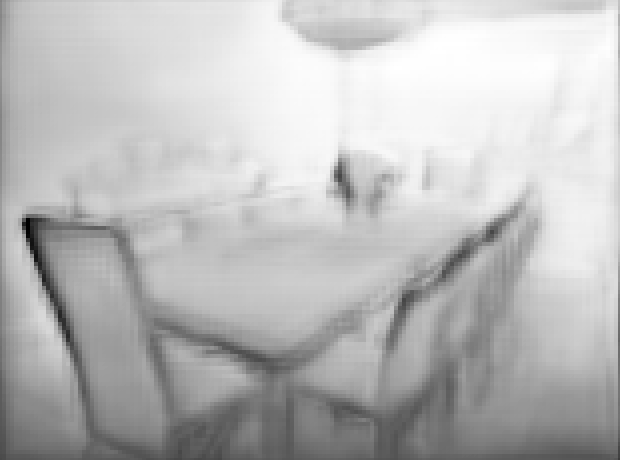}
	\end{subfigure}
	\begin{subfigure}{.3\linewidth}
		\centering
		\includegraphics[width=\linewidth]{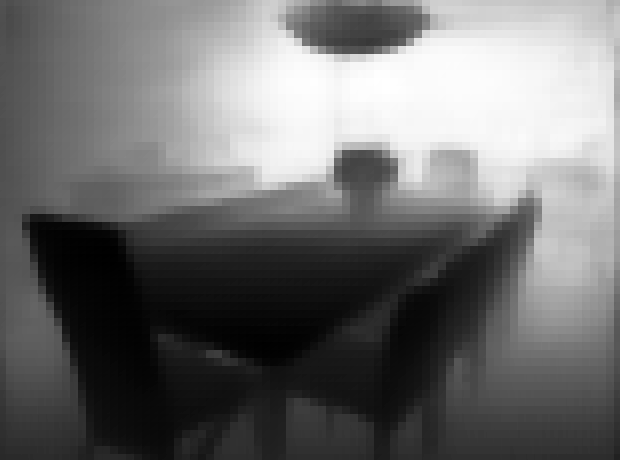}
	\end{subfigure}
	\caption{Example outputs from the proposed network. Top: from left to right, input image, predicted depth map, and the ground truth. Bottom: from left to right, outputs from the proposed local planar guidance layers having input feature resolutions of 1/2, 1/4, 1/8 to the input image, respectively.}
	\label{fg:mde_example}
\end{figure}

In the meantime, recent applications based on DCNNs are commonly composed in two parts: encoder for dense feature extraction and decoder for the desired prediction.
As a dense feature extractor, very powerful deep networks such as VGG \cite{simonyan2014very}, ResNet \cite{he2016deep} or DenseNet \cite{huang2017densely} are usually adopted.
In these networks, repeated strided convolution and spatial pooling layers lower the spatial resolution of transitional outputs, which can be a bottleneck to obtain desired dense predictions.
Therefore, a number of techniques, for example, multi-scale networks \cite{liu2016learning,eigen2015predicting}, skip connections \cite{godard2017unsupervised,xie2016deep3d} or multi-layer deconvolutional networks \cite{laina2016deeper,garg2016unsupervised,kuznietsov2017semi} are applied to consolidate feature maps from higher resolutions.
Recently, atrous spatial pyramid pooling (ASPP) \cite{chen2018deeplab} has been introduced for image semantic segmentation, which can capture large scale variations in observation by applying sparse convolutions with various dilation rates.
Since the dilated convolution allows larger receptive field size, recent works in semantic segmentation \cite{chen2018deeplab,yang2018denseaspp} or depth estimation \cite{fu2018deep} do not fully reduce the receptive field size by removing last few pooling layers and reconfigure the network with atrous convolutions to reuse pre-trained weights.
Consequently, their methods have larger dense features (1/8 of input spatial resolution whereas 1/32 or 1/64 in the original base networks) and perform almost all of the decoding process on that resolution followed by a simple upsampling to recover the original spatial resolution.

\begin{figure*}[t!]
	\centering	
	\includegraphics[width=\textwidth]{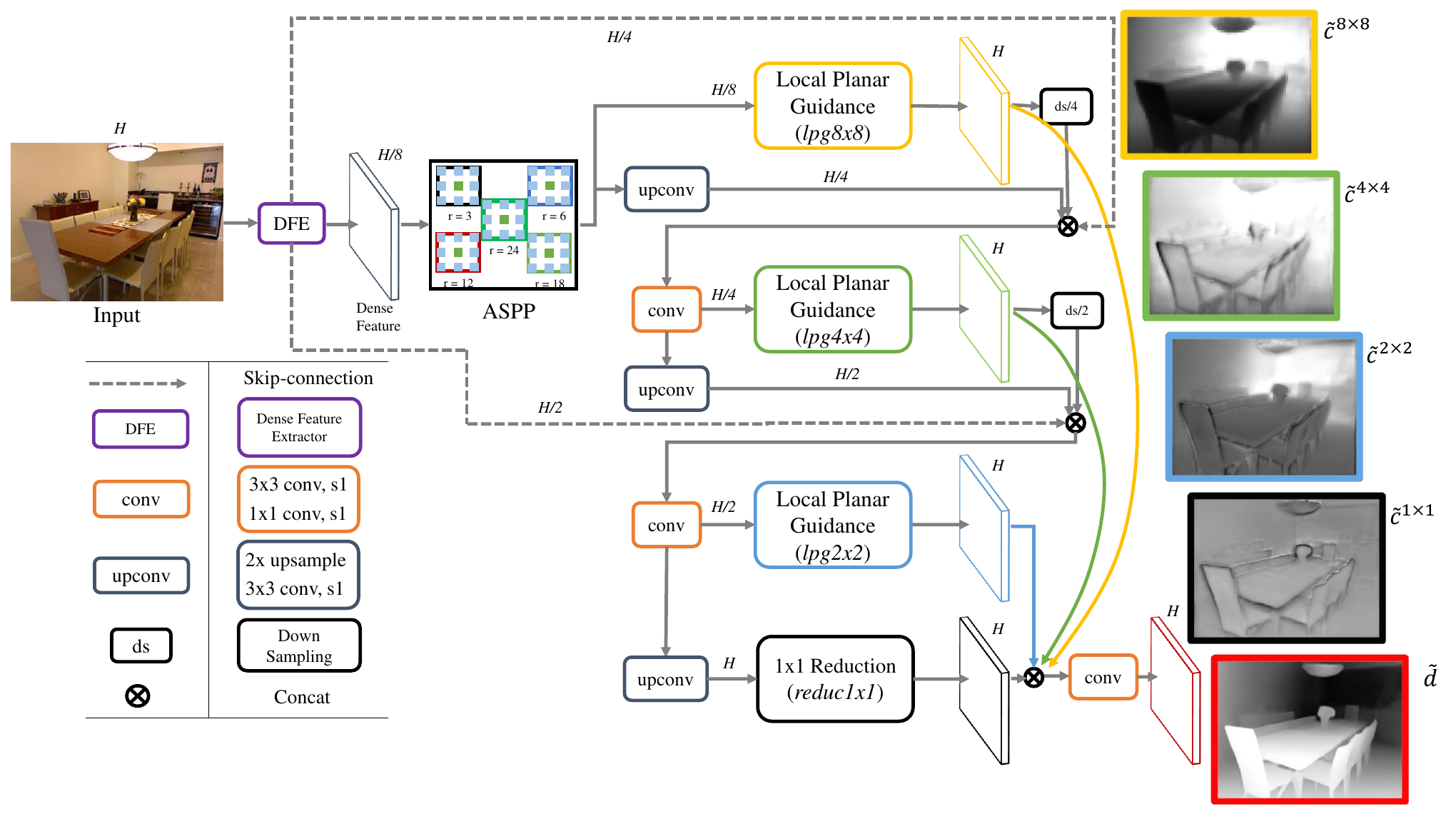}
	\caption{Overview of the proposed network architecture.
		The network is composed of dense feature extractor (the base network), contextual information extractor (ASPP), local planar guidance layers and their dense connection for final depth estimation.
		Note that the outputs from the local planar guidance layers have the full spatial resolution $H$ enabling shortcuts inside the decoding phase.
		We also use skip-connections from the base network to link with internal outputs in the decoding phase with corresponding spatial resolutions.}
	\label{fg:architecture}
\end{figure*}

To define explicit relation in recovering back to the full resolution, we propose a network architecture that utilizes novel local planar guidance layers located at multiple stages in the decoding phase.
Specifically, based on an encoding-decoding scheme, at each decoding stage, which has spatial resolutions of 1/8, 1/4, and 1/2, we place a layer that effectively guides input feature maps to the desired depth with local planar assumption.
Then, we combine the outputs to predict depth in full resolution.
This differs from multi-scale network \cite{eigen2015predicting,eigen2014depth} or image pyramid \cite{godard2017unsupervised} approaches in two aspects.
First, the outputs from the proposed layers are not treated as separated global depth estimation in corresponding resolutions.
Instead, we let the layers to learn 4-dimensional plane coefficients and use them together to reconstruct depth estimations in the full resolution for the final estimation.
Second, as a consequence of the nonlinear combination, individual spatial cells in each resolution are distinctively trained while the training progresses.
We can see example outputs from the proposed layers in Figures \ref{fg:mde_example} and \ref{fg:lpg_examples}.
Experiments on the challenging NYU Depth V2 dataset \cite{silberman2012indoor} and KITTI dataset \cite{geiger2013vision} demonstrate that the proposed method achieves state-of-the-art results.

The rest of the paper is organized as follows.
After a concise survey of related works in Section \ref{sec:related_work}, we present in detail the proposed method in Section \ref{sec:method}.
Then, in Section \ref{sec:experiments}, we provide results on two challenging benchmarks comparing with state-of-the-art works, and using various base networks as an encoder for the proposed network, we see how the performance varies along with each base network.
In Section \ref{sec:experiments}, we also provide an ablation study to validate the effectiveness of the proposed method.
We conclude the paper in Section \ref{sec:conclusion}.

\section{Related Work}
\label{sec:related_work}

\subsection{Supervised Monocular Depth Estimation}
In monocular depth estimation, supervised approaches take a single image and use depth data measured with range sensors such as RGB-D cameras or multi-channel laser scanners as ground truth for supervision in training.
Saxena et al. \cite{saxena2006learning} propose a learning-based approach to get a functional mapping from visual cues to depth via Markov random field, and extend it to a patch-based model that first over-segments the input image and learns 3D orientation as well as the location of local planes that are well explained by each patch \cite{saxena2009make3d}.
Eigen et al. \cite{eigen2015predicting} introduce a multi-scale convolutional architecture that learns coarse global depth predictions on one network and progressively refine them using another network.
Unlike the previous works in single image depth estimation, their network can learn representations from raw pixels without handcrafted features such as contours, super-pixels, or low-level segmentation.
Several works follow the success of this approach by incorporating strong scene priors for surface normal estimation \cite{wang2015designing}, using conditional random fields to improve accuracy \cite{li2015depth,knobelreiter2017end,schwing2015fully} or changing the learning problem from regression to classification \cite{cao2017estimating}.
A recent supervised approach from Fu et al. \cite{fu2018deep} achieves the state-of-the-art result by also taking advantage of changing the regression problem to quantized ordinal regression.
Xu et al. \cite{xu2017multi} propose an architecture that exploits multi-scale estimations derived from inner layers by fusing them within a CRF framework.
Gan et al. \cite{gan2018monocular} propose to explicitly model the relationships between different image locations with an affinity layer.
Most recently, Yin et al. \cite{yin2019enforcing} introduce a method using virtual normal directions that are determined by randomly chosen three points in the reconstructed 3D space as geometric constraints.

\subsection{Semi-Supervised Monocular Depth Estimation}
There are also attempts to train a depth estimation network in a semi-supervised or weakly supervised fashion.
Chen et al. \cite{chen2016single} propose a new approach that uses information of relative depth and depth ranking loss function to learn depth predictions in unconstrained images.
Recently, to overcome the difficulty in getting high-quality depth data, Kuznietsov et al. \cite{kuznietsov2017semi} introduce a semi-supervised method to train the network using both sparse LiDAR depth data for direct supervision and image alignment loss as a secondary training objective.

\subsection{Self-Supervised Monocular Depth Estimation}
The self-supervised approach refers to a method that requires only rectified stereo image pairs to train the depth estimation network.
Garg et al. \cite{garg2016unsupervised} and Godard et al. \cite{godard2017unsupervised} propose \textit{self-supervised} learning methods that smartly cast the problem from direct depth estimation to image reconstruction.
Specifically, with a rectified stereo image pair, their networks try to synthesize one view from the other with estimated disparities and define the error between both as the reconstruction loss for the main training objective.
In this way, because learning requires only well rectified, synchronized stereo pairs instead of the ground truth depth data well associated with the corresponding RGB images, it greatly reduces the effort to acquire datasets for new categories of scenes or environments.
However, there is some accuracy gap when compared to the current best supervised approach \cite{yin2019enforcing}.
Garg et al. \cite{garg2016unsupervised} introduce an encoder-decoder architecture and to train the network using photometric reconstruction error. 
Xie et al. \cite{xie2016deep3d} propose a network that also synthesizes one view from the other, and by using the reconstruction error, they produce a probability distribution of possible disparities for each pixel.
Godard et al. \cite{godard2017unsupervised} finally propose a network architecture fully differentiable thus can perform end-to-end training.
They also present a novel left-right consistency loss that improves training and predictions of the network.
Most recently, Godard et al. \cite{godard2019digging} propose a simple but effective architecture benefiting from associated design choices such as a robust reprojection loss, multi-scale sampling, and an auto-masking loss.

\subsection{Video-Based Monocular Depth Estimation}
There are also approaches using sequential data to perform the monocular depth estimation.
Yin et al. \cite{yin2018geonet} propose an architecture consisting of two generative sub-networks that are jointly trained by adversarial learning for disparity map estimation organized in a cycle to provide mutual constraints.
Mahjourian et al. \cite{mahjourian2018unsupervised} presents an approach that explicitly considers the inferred 3D geometry of the whole scene, and enforce consistency of the estimated 3D point clouds and ego-motion across consecutive frames.
Wang et al. \cite{wang2018learning} adopt a differentiable pose predictor and train a monocular depth estimation network in an end-to-end fashion while benefiting from the pose predictor.

\section{Method}
\label{sec:method}
In this section, we describe the proposed monocular depth estimation network with a novel local planar guidance layer located on multiple stages in the decoding phase.

\subsection{Network Architecture}
As it can be seen from Figure \ref{fg:architecture}, we follow an encoding-decoding scheme that reduces feature map resolution to $H/8$ then recovers the original resolution $H$ for dense prediction. 
After the backbone network that we use as a dense feature extractor which produces an $H/8$ feature map, we place a denser version \cite{yang2018denseaspp} of atrous spatial pyramid pooling layer \cite{chen2018deeplab} as our contextual information extractor with various dilation rates $r\in\{3,6,12,18,24\}$.
Then, at each stage in the decoding phase, where internal outputs are recovered to the full spatial resolution with a factor of 2, we employ the proposed local planar guidance (LPG) layer to locate geometric guidance to the desired depth estimation.
We also place a 1$\times$1 reduction layer to get the finest estimation $\tilde{c}^{1\times 1} \in R^{H\times W\times 1}$ after the last \textit{upconv} layer.
Finally, outputs from the proposed layers (\textit{i.e.,} $\tilde{c}^{k\times k}$) and $\tilde{c}^{1\times 1}$ are concatenated and fed into the final convolutional layer to get the depth estimation $\tilde{d}$.

\begin{figure}[t!]
	\centering
	\footnotesize
	\includegraphics[width=\linewidth]{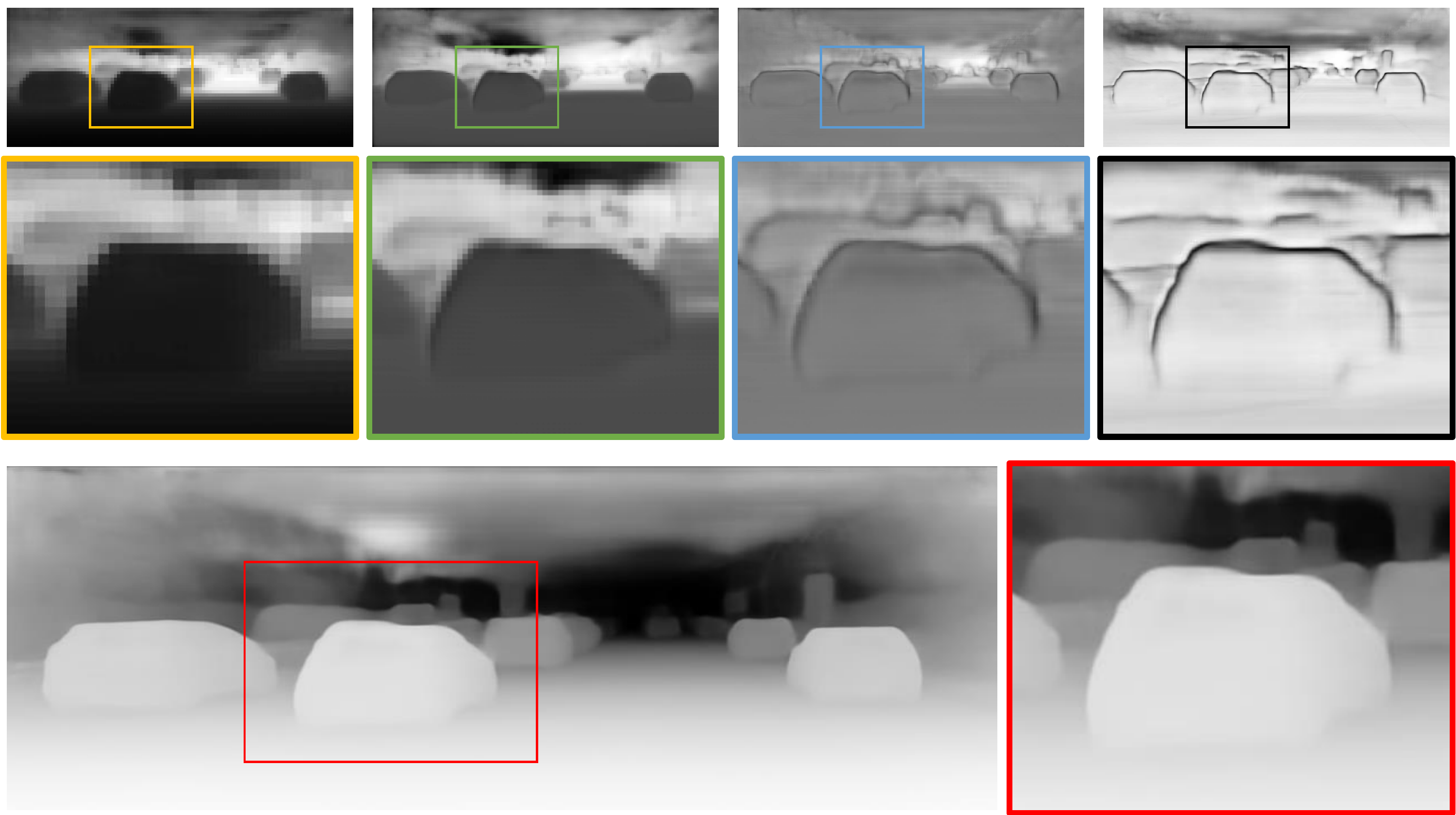}
	\caption{Examples showing behavior of the proposed network. Top and middle rows show $\tilde{c}^{k\times k}$ and their focused views. Bottom row shows the final depth estimation result with a focused view. The overestimated boundaries of the vehicle from lpg8x8 (yellow rectangle) and lpg4x4 (green rectangle) are compensated by the outputs from lpg2x2 (blue rectangle) and reduc1x1 (black rectangle) (\textit{i.e,} $\tilde{c}^{2\times 2}$ and $\tilde{c}^{1\times 1}$), resulting the clear boundary in the final estimation.}
	\label{fg:lpg_examples}
	\vspace{-0.4cm}
\end{figure}

\subsection{Multi-Scale Local Planar Guidance}
Our key idea in this work is to define direct and explicit relations between internal features and the final output in an effective manner.
Unlike the existing methods that recover back to the original resolution using simple nearest neighbor upsampling layers and skip connections from encoding stages, we place novel local planar guidance layers which guide features to the full resolution with the local planar assumption and use them together to get the final depth estimation $\tilde{d}$.
As can be seen from Figure \ref{fg:architecture}, since the proposed layer recovers given an internal feature map to the full resolution $H$, it can be used as a skip connection inside the decoding phase allowing direct relations between internal features and the final prediction.
Specifically, given a feature map having spatial resolution $H/k$, the proposed layers estimate for each spatial cell a 4D plane coefficients that fit a locally defined $k\times k$ patch on the full resolution $H$, and they are concatenated together for the prediction through the final convolutional layers.

Please note that the proposed LPG layer is not designed to directly estimate global depth values on the corresponding scale because the training loss is only defined in terms of the final depth estimation (provided in Section \ref{sec:training_loss}).
Together with outputs from the other LPG layers and \textit{reduc1x1}, each output is interpreted to the global depth by contributing as a part of the nonlinear combination through the final convolutional layers.
Therefore, they can have distinct ranges, learned as a base or precise relative compensation from the base at a spatial location, as shown in Figures \ref{fg:mde_example} and \ref{fg:lpg_examples}.

\begin{figure}[t]
	\centering
	\footnotesize
	\includegraphics[width=0.9\linewidth]{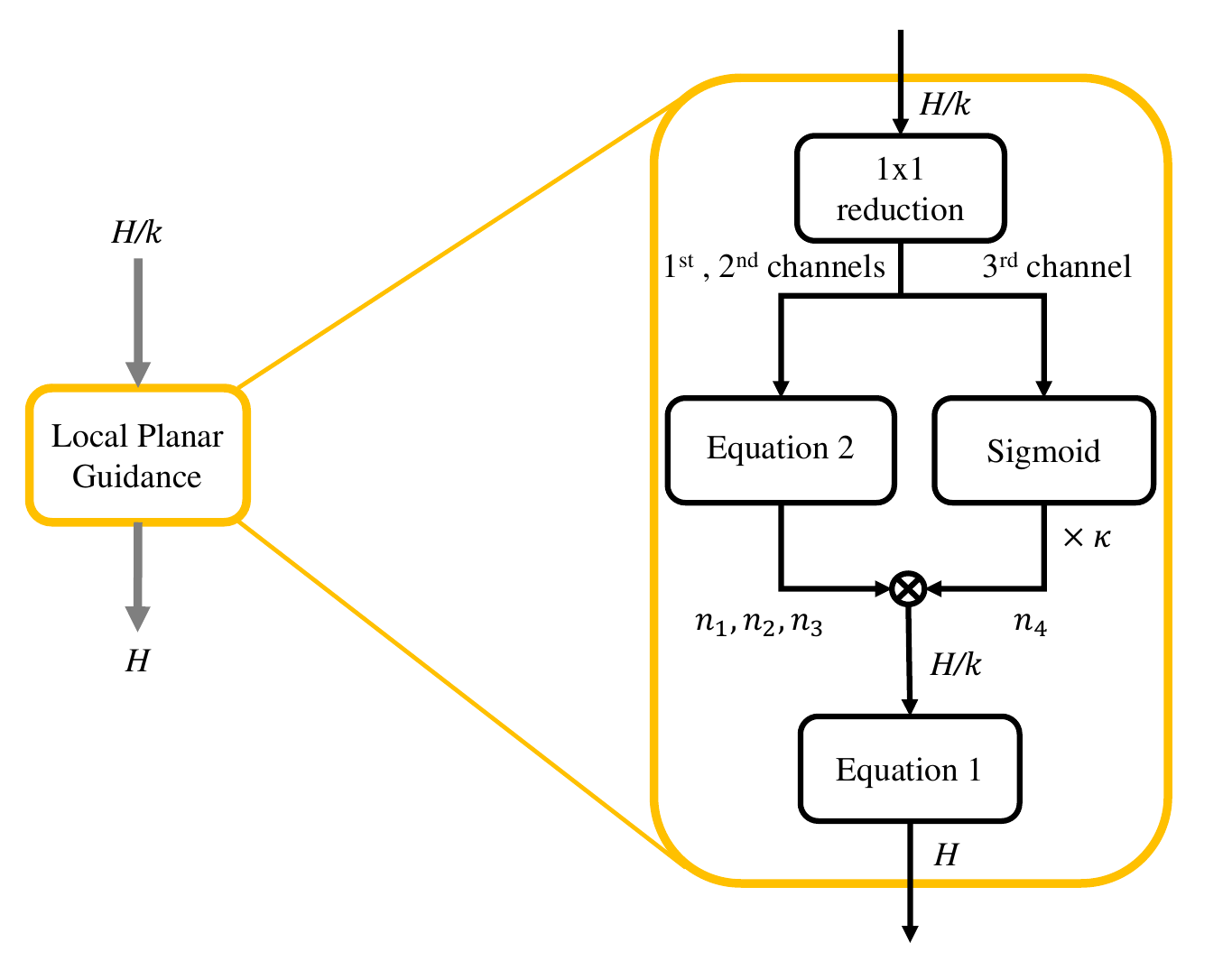}
	\caption{The local planar guidance layer. We use a stack of $1\times1$ convolutions to get 4D coefficients estimations. (\textit{i.e.,} $H/k\times H/k\times 4$). Then the channels are split to pass through two different activation mechanisms to ensure plane coefficients' constraint. Finally, they are fed into the planar guidance module to compute locally-defined relative depth estimations.}
	\label{fg:local_planar_guidance}
\end{figure}

Here, we use the local planar assumption because, for a $k\times k$ region, it enables an efficient reconstruction with only four parameters.
If we adopt typical \textit{upconvs} for the reconstruction, the layers should be learned to have $k^2$ values properly instead of four.
Therefore, we can expect that our strategy can be more effective because conventional upsampling would not give details on enlarged resolutions, while the local linear assumption can provide effective guidance.

To guide features with the local planar assumption, we convert each estimated 4D plane coefficients to $k\times k$ local depth cues using \textit{ray-plane intersection}:
\begin{equation}
	\tilde{c}_i=\frac{n_4}{n_1u_i+n_2v_i+n_3},
	\label{eq:compute_depth}
\end{equation}
where $n=(n_1, n_2, n_3, n_4)$ denotes the estimated plane coefficients, $(u_i,v_i)$ are $k\times k$ patch-wise normalized coordinates of pixel $i$.

\begin{table}[t]
	\centering
	\resizebox{1\columnwidth}{!}{%
		\begin{tabular}{r|ccc|ccc}
			\hline
			\multicolumn{1}{c|}{Method}       & $\delta<1.25$ & $\delta<1.25^2$ & $\delta<1.25^3$ & AbsRel & RMSE  & log10 \\ \hline \hline
			Saxena et al. \cite{saxena2009make3d} & 0.447 & 0.745 & 0.897 & 0.349 & 1.214 & \multicolumn{1}{c}{-} \\
			Wang et al. \cite{wang2015towards}  & 0.605 & 0.890 & 0.970 & 0.220 & 0.824 & \multicolumn{1}{c}{-} \\
			Liu et al. \cite{liu2016learning}   & 0.650 & 0.906 & 0.976 & 0.213 & 0.759 & 0.087                  \\
			Eigen et al. \cite{eigen2015predicting} & 0.769 & 0.950 & 0.988 & 0.158 & 0.641 & \multicolumn{1}{c}{-} \\
			Chakrabarti et al. \cite{chakrabarti2016depth} & 0.806 & 0.958 & 0.987 & 0.149 & 0.620 & \multicolumn{1}{c}{-} \\
			Li et al. \cite{li2017two}   & 0.789 & 0.955 & 0.988 & 0.152 & 0.611 & 0.064                  \\
			Laina et al. \cite{laina2016deeper} & 0.811 & 0.953 & 0.988 & 0.127 & 0.573 & 0.055                  \\
			Xu et al. \cite{xu2017multi}   & 0.811 & 0.954 & 0.987 & 0.121 & 0.586 & 0.052                  \\
			Lee at al. \cite{lee2018single}   & 0.815 & 0.963 & 0.991 & 0.139 & 0.572 & \multicolumn{1}{c}{-} \\
			Fu et al. \cite{fu2018deep}    & 0.828 & 0.965 & 0.992 & 0.115 & 0.509 & 0.051 \\
			Qi et al. \cite{qi2018geonet}    & 0.834 & 0.960 & 0.990 & 0.128 & 0.569 & 0.057 \\ 
			Yin et al. \cite{yin2019enforcing} & 0.875 & 0.976 & 0.994 & \bf{0.108} & 0.416 & 0.048 \\ \hline
			Ours-ResNet   & 0.871 & 0.977 & 0.995 & 0.113 & 0.407 & 0.049 \\
			Ours-ResNext  & 0.880 & 0.977 & 0.994 & 0.111 & 0.410 & 0.048 \\
			Ours-DenseNet & \bf{0.885} & \bf{0.978} & \bf{0.994} & 0.110 & \bf{0.392} & \bf{0.047} \\ \hline
		\end{tabular}
	}
	\caption{Evaluation results on NYU Depth v2. Ours outperforms previous works with a significant margin in all measures except only from AbsRel.}
	\label{tb:nyu}	
	\vspace{-0.5cm}
\end{table}

Figure \ref{fg:local_planar_guidance} shows the detail of the proposed layer.
Through a stack of $1\times1$ convolutions which repeatedly reduce the number of channels by a factor of 2 until it reaches 3, we get a $H/k\times H/k \times 3$ feature map if we assume a square input.
Then, we pass the feature map through two different ways to get local plane coefficient estimations: one way is a conversion to a unit normal vector $(n_1, n_2, n_3)$, and the other is a sigmoid function defining the perpendicular distance $n_4$ between the plane and origin.
After the sigmoid function we multiply the output with the maximum distance $\kappa$ to get real depth values.
Because a unit normal vector has only two degrees of freedom (\textit{i.e.,} polar and azimuthal angles $\theta, \phi$ from predefined axes), we regard the first two channels of the given feature map as the angles and convert them to unit normal vectors using following equations.
\begin{equation}
	n_1 = \sin(\theta)\cos(\phi),
	n_2 = \sin(\theta)\sin(\phi),
	n_3 = \cos(\theta).
\end{equation}
Finally, they are concatenated again and used for estimation of $\tilde{c}^{k\times k}$ using Equation \ref{eq:compute_depth}.

We design the local depth cue as an additive depth defined in local regions (\textit{i.e.,} $k\times k$ patches).
Since features at the same spatial location in different stages are used together to predict the final depth, for efficient representation, we expect that global shapes would be learned at coarser scales while local details at finer scales.
Also, they can interact with each other to compensate for erroneous estimations.
We can represent the behavior of the last convolutional layer as follows.
\begin{equation}
	\tilde{d}=f\left(W_1\tilde{c}^{1\times 1}+W_2\tilde{c}^{2\times 2}+W_3\tilde{c}^{4\times 4}+W_4\tilde{c}^{8\times 8}\right),
\end{equation}
where $f$ is an activation function, $W_j, j \in \{1,2,3,4\} $ denotes a corresponding linear transform representing the convolution.
Please note that the proposed network learns on multiple scales, and by defining the training loss only in terms of the final estimation, $\tilde{d}$, we do not enforce parameters for each scale learns with the constant contribution.
Therefore, in training, details for regions with sharp curvatures would be learned at finer scales while major structures at coarser scales.
Also, there is the last chance in $\tilde{c}^{1\times 1}$ to recover broken assumptions in the upsampled estimations ($\tilde{c}^{k\times k}, k\in\{2,4,8\}$).
From Figure \ref{fg:lpg_examples}, we can see small details behind the focused vehicle from blue- and black-boxed figures which demonstrate $\tilde{c}^{2\times 2}$ and $\tilde{c}^{1\times 1}$, respectively, while they are missing in the coarser scales, $\tilde{c}^{8\times 8}$ and $\tilde{c}^{4\times 4}$.
Also, there are thick black estimations in $\tilde{c}^{1\times 1}$ and $\tilde{c}^{2\times 2}$ on the boundary of the vehicle compensating the over-estimations in $\tilde{c}^{8\times 8}$ and $\tilde{c}^{4\times 4}$.
More examples are provided in the supplementary material.

\begin{table*}[t!]
	\centering
	\resizebox{2\columnwidth}{!}{%
		\begin{tabular}{r|l|ccc|cccc}
			\hline
			\multicolumn{1}{c|}{\multirow{2}{*}{Method}} & \multicolumn{1}{c|}{\multirow{2}{*}{cap}} & \multicolumn{3}{c|}{\textit{higher is better}} & \multicolumn{4}{c}{\textit{lower is better}} \\ \cline{3-9}
			& \multicolumn{1}{c|}{} & $\delta<1.25$ & $\delta<1.25^2$ & $\delta<1.25^3$ & Abs Rel & Sq Rel & RMSE & RMSE \textit{log} \\ \hline \hline
			Saxena et al. \cite{saxena2009make3d}              & 0-80m & 0.601 & 0.820 & 0.926 & 0.280 & 3.012 & 8.734 & 0.361 \\
			Eigen et al. \cite{eigen2014depth}                 & 0-80m & 0.702 & 0.898 & 0.967 & 0.203 & 1.548 & 6.307 & 0.282 \\
			Liu et al. \cite{liu2016learning}                  & 0-80m & 0.680 & 0.898 & 0.967 & 0.201 & 1.584 & 6.471 & 0.273 \\
			Godard et al. (CS+K) \cite{godard2017unsupervised} & 0-80m & 0.861 & 0.949 & 0.976 & 0.114 & 0.898 & 4.935 & 0.206 \\
			Kuznietsov et al. \cite{kuznietsov2017semi}        & 0-80m & 0.862 & 0.960 & 0.986 & 0.113 & 0.741 & 4.621 & 0.189 \\
			Godard et al. (CS+K) \cite{godard2017unsupervised} & 0-80m & 0.861 & 0.949 & 0.976 & 0.114 & 0.898 & 4.935 & 0.206 \\
			Gan et al. \cite{gan2018monocular}                 & 0-80m & 0.890 & 0.964 & 0.985 & 0.098 & 0.666 & 3.933 & 0.173 \\
			Fu et al. \cite{fu2018deep}                        & 0-80m & 0.932 & 0.984 & 0.994 & 0.072 & 0.307 & \textbf{2.727} & 0.120 \\ 
			Yin et al. \cite{yin2019enforcing}                 & 0-80m & 0.938 & 0.990 & \bf{0.998} & 0.072 &   -   & 3.258 & 0.117 \\ \hline
			Ours-ResNet                                        & 0-80m & 0.954 & 0.992 & \bf{0.998} & 0.061 & 0.261 & 2.834 & 0.099 \\
			Ours-DenseNet                                      & 0-80m & 0.955 & \bf{0.993} & \bf{0.998} & 0.060 & 0.249 & 2.798 & \bf{0.096} \\
			Ours-ResNext                                       & 0-80m & \bf{0.956} & \bf{0.993} & \bf{0.998} & \bf{0.059} & \bf{0.245} & 2.756 & \bf{0.096} \\ \hline \hline
			Garg et al. \cite{garg2016unsupervised}            & 0-50m & 0.740 & 0.904 & 0.962 & 0.169 & 1.080 & 5.104 & 0.273 \\
			Godard et al. (CS+K) \cite{godard2017unsupervised} & 0-50m & 0.873 & 0.954 & 0.979 & 0.108 & 0.657 & 3.729 & 0.194 \\
			Kuznietsov et al. \cite{kuznietsov2017semi}        & 0-50m & 0.875 & 0.964 & 0.988 & 0.108 & 0.595 & 3.518 & 0.179 \\
			Gan et al. \cite{gan2018monocular}                 & 0-50m & 0.898 & 0.967 & 0.986 & 0.094 & 0.552 & 3.133 & 0.165 \\
			Fu et al. \cite{fu2018deep}                        & 0-50m & 0.936 & 0.985 & 0.995 & 0.071 & 0.268 & 2.271 & 0.116 \\ \hline
			Ours-ResNet                                        & 0-50m & 0.962 & 0.994 & \bf{0.999} & 0.058 & 0.183 & 1.995 & 0.090 \\			
			Ours-DenseNet                                      & 0-50m & \bf{0.964} & \bf{0.995} & \bf{0.999} & 0.057 & 0.175 & 1.949 & 0.088 \\
			Ours-ResNext                                       & 0-50m & \bf{0.964} & 0.994 & \bf{0.999} & \bf{0.056} & \bf{0.169} & \bf{1.925} & \bf{0.087} \\ \hline
		\end{tabular}
	}
	\caption{Performance on KITTI Eigen split.
	(CS+K) denotes a model pre-trained on Cityscapes dataset \cite{cordts2016cityscapes} and fine-tuned on KITTI.
	}
	\label{tb:result_kitti_eigen}
\end{table*}

\subsection{Training Loss}
\label{sec:training_loss}
In \cite{eigen2014depth}, Eigen et al introduce a scale-invariant error and inspired from it they use a following training loss:
\begin{equation}
	D(g) = \frac{1}{T}\sum_i g_i^2 - \frac{\lambda}{T^2}\left(\sum_i g_i\right)^2,
	\label{eq:siloss}
\end{equation}
where $g_i=\log \tilde{d}_i - \log d_i$ with the ground truth depth $d_i$, $\lambda=0.5$ and $T$ denotes the number of pixels having valid ground truth values.
By rewriting above equation,
\begin{align*}
	D(g) = \frac{1}{T}\sum_i g_i^2 - \left(\frac{1}{T} \sum_i g_i\right)^2 + (1-\lambda)\left(\frac{1}{T} \sum_i g_i\right)^2,
\end{align*}
we can see that it is a sum of the variance and a weighted squared mean of the error in log space.
Therefore, setting a higher $\lambda$ enforces more focusing on minimizing the variance of the error, and we use $\lambda=0.85$ in this work.
Also, we observe that properly scaling the range of the loss function improves convergence and the final training result.
Finally, we define our training loss $L$ as follows:
\begin{equation}
	L = \alpha \sqrt{D(g)},
	\label{eq:loss}
\end{equation}
where $\alpha$ is a constant we set to 10 for all experiments.

\begin{table}[t]
	\begin{center}
		\centering		
		\resizebox{\columnwidth}{!}{%
			\begin{tabular}{r|c|c|c|c}
				\hline
				\multicolumn{1}{c|}{Method} & SILog & sqErrorRel & absErrorRel & iRMSE \\ \hline \hline
				Yin et al. \cite{yin2019enforcing} & 12.65 & 2.46 & 10.15 & 13.02 \\
				Diaz et al. \cite{diaz2019soft}    & 12.39 & 2.49 & 10.10 & 13.48 \\
				Fu et al. \cite{fu2018deep}        & 11.77 & 2.23 & \bf{8.78} & 12.98 \\ \hline
				Ours                               & \bf{11.67} & \bf{2.21} & 9.04 & \bf{12.23} \\ \hline
			\end{tabular}
		}	
	\end{center}
	\caption{Result on the online KITTI evaluation server.}
	\label{tb:kitti_online_benchmark}
\end{table}

\section{Experiments}
\label{sec:experiments}
To verify the effectiveness of our approach, we provide experimental results from challenging benchmarks with various settings.
After presenting the implementation details of our method, we provide experimental results on two challenging benchmarks covering both indoor and outdoor environments.
We also provide scores on the online KITTI evaluation server comparing with published works.
Then, we provide an ablation study to discuss a detailed analysis of the proposed core factors, and some qualitative results to demonstrate our approach comparing with competitors.

\begin{table*}[t]
	\centering
	\normalsize
	\resizebox{2\columnwidth}{!}{%
		\begin{tabular}{l|c|ccc|ccccc}
			\hline
			\multicolumn{1}{c|}{\multirow{2}{*}{Variant}} & \multicolumn{1}{c|}{\multirow{2}{*}{\# Params}} & \multicolumn{3}{c|}{\textit{higher is better}} & \multicolumn{5}{c}{\textit{lower is better}} \\ \cline{3-10} 
			\multicolumn{1}{c|}{} &  \multicolumn{1}{c|}{} & $\delta<1.25$ & $\delta<1.25^2$ & $\delta<1.25^3$ & Abs Rel & Sq Rel & RMSE & RMSE \textit{log} & log10 \\ \hline
			Baseline             & 63.0M & 0.815 & 0.958 & 0.990 & 0.142 & 0.084 & 0.587 & 0.169 & 0.053 \\
			Baseline + A         & 67.4M & 0.827 & 0.964 & 0.992 & 0.141 & 0.081 & 0.577 & 0.166 & 0.051 \\
			Baseline + A + U     & 68.4M & 0.845 & 0.967 & 0.992 & 0.134 & 0.076 & 0.513 & 0.161 & 0.051 \\
			Baseline + A + U + L & 68.5M & 0.863 & 0.974 & 0.994 & 0.119 & 0.072 & 0.421 & 0.149 & 0.050 \\ \hline
			Ours-ResNet          & 68.5M & 0.871 & 0.975 & 0.995 & 0.113 & 0.068 & 0.407 & 0.148 & 0.049 \\
			Ours-DenseNet        & 47.0M & \bf{0.885} & \bf{0.978} & \bf{0.994} & \bf{0.110} & \bf{0.066} & \bf{0.392} & \bf{0.142} & \bf{0.047} \\ \hline
		\end{tabular}
	}
	\caption{Result from the ablation study using the NYU Depth V2 dataset. Baseline: a network composed of only the dense feature extractor and direct estimation from it followed by an upsampling with a factor of 8, A: ASPP module attached after the dense feature extractor, U: using \textit{upconv} layers in Figure \ref{fg:architecture}, L: the proposed local planar guidance layers. All variants are trained using \textit{ResNet-101} as the base network and Equation \ref{eq:siloss} with $\lambda=0.5$ as the training loss. `Ours-ResNet' and `Ours-DenseNet' use the training loss given in Equation \ref{eq:loss}.}
	\label{tb:ablation_study}	
\end{table*}

\subsection{Implementation Details}
We implement the proposed network using the open deep learning framework \textit{PyTorch} \cite{NEURIPS2019_9015}.
For training, we use Adam optimizer \cite{kingma2014adam} with $\beta_1=0.9, \beta_2=0.999$ and $\epsilon=10^{-6}$, learning is scheduled via polynomial decay from base learning rate $10^{-4}$ with power $p=0.9$.
The total number of epochs is set to 50 with batch size 16 on a desktop equipped with four NVIDIA 1080ti GPUs for all experiments in this work.

As the backbone network for dense feature extraction, we use ResNet-101 \cite{he2016deep}, ResNext-101 \cite{xie2017cvpr} and DenseNet-161 \cite{huang2017densely} with pretrained weights trained for image classification using ILSVRC dataset \cite{russakovsky2015imagenet}.
Because weights at early convolutions are known to be well trained for primitive visual features, in the base networks, we fix the first two convolutional layers as well as batch normalization parameters in our training.
Following \cite{godard2017unsupervised}, we use exponential linear units \cite{clevert2015fast} as an activation function, and \textit{upconv} uses the nearest neighbor upsampling followed by a $3\times3$ convolution layer \cite{odena2016deconvolution}.

To avoid over-fitting, we augment images before input to the network using random horizontal flipping as well as random contrast, brightness, and color adjustment in a range of [0.9, 1.1], with 50\% of chance.
We also use a random rotation of the input images in ranges of $\left[-1, 1\right]$ and $\left[-2.5, 2.5\right]$ degrees for KITTI and NYU datasets, respectively.
We train our network on a random crop of size $352\times704$ for KITTI and $416\times544$ for NYU Depth V2 datasets.

\subsection{NYU Depth V2 Dataset}
The NYU Depth V2 dataset \cite{silberman2012indoor} contains 120K RGB and depth pairs having a size of $480\times640$ acquired as video sequences using a Microsoft Kinect from 464 indoor scenes.
We follow the official train/test split as previous works, using 249 scenes for training and 215 scenes (654 images) for testing.
From the total 120K image-depth pairs, due to asynchronous capturing rates between RGB images and depth maps, we associate and sample them using timestamps by even-spacing in time, resulting in 24231 image-depth pairs for the training set.
Using raw depth images and camera projections provided by the dataset, we align the image-depth pairs for accurate pixel registrations.
We use $\kappa=10$ for this dataset.

\subsection{KITTI Dataset}
KITTI provides the dataset \cite{geiger2013vision} with 61 scenes from ``city", ``residential", ``road" and ``campus" categories.
Because existing works commonly use a split proposed by Eigen et al. \cite{eigen2014depth} for the training and testing, we also follow it to compare with those works.
Therefore, 697 images covering a total of 29 scenes are used for evaluation, and the remaining 32 scenes of 23,488 images are used for the training.
We use $\kappa=80$ for this dataset.

\subsection{Evaluation Result}
For evaluation, we use following metrics used by previous works:\\
$\small\text{Threshold}: \text{\% of } \tilde{d}_i \text{ s.t.} \max(\frac{\tilde{d}_i}{d_i}, \frac{d_i}{\tilde{d}_i})=\delta<thr,$
$\small\text{Abs Rel}: \frac{1}{|T|}\sum_{\tilde{d}\in T}|\tilde{d}-d|/d,$ \\
$\small\text{Sq Rel}: \frac{1}{|T|}\sum_{\tilde{d}\in T}||\tilde{d}-d||^2/d,$
$\small\text{RMSE}: \sqrt{\frac{1}{|T|}\sum_{\tilde{d}\in T}||\tilde{d}-d||^2},$ \\
$\small\text{log10}: \frac{1}{|T|}\sum_{\tilde{d}\in T}|\log_{10} \tilde{d}-\log_{10} d|,$
$\small\text{RMSE\textit{log}}: \sqrt{\frac{1}{|T|}\sum_{\tilde{d}\in T}||\log \tilde{d}-\log d||^2},$
where $T$ denotes a collection of pixels that the ground truth values are available.

Using NYU Depth V2 dataset, the experimental results given in Table \ref{tb:nyu} show that Ours-DenseNet achieves the state-of-the-art result with a significant margin in both of the inlier measures (\textit{i.e.,} $\delta<thr$) and accuracy metrics (\textit{i.e.,} AbsRel, log10) except only RMSE. Our ResNet-based model also outperforms the method from Yin et al \cite{yin2019enforcing} with a significant margin which has the same backbone network, ResNext-101 \cite{xie2017cvpr}.

In the evaluation using the KITTI dataset provided in Table \ref{tb:result_kitti_eigen}, ours outperforms all existing works with a significant margin.
Please note that our ResNet-based model already outperforms the methods from Fu et al \cite{fu2018deep} and Yin et al \cite{yin2019enforcing} which use ResNet-101 and ResNext-101 \cite{xie2017cvpr} as their backbone network, respectively.
Only the root mean squared errors (RMSE) from ours are behind that of Fu et al.'s in the capturing range 0-80m.
However, in the capturing range 0-50m, ours-ResNet achieves more than 10\% improvement in RMSE from the result of Fu et al.
Also, the proposed method achieves notable improvements in the inlier metrics (\textit{i.e.,} $\delta < thres$), meaning more number of correctly estimated pixels as it can be seen from Figures \ref{fg:kitti_qualitative_result} and \ref{fg:nyu_qualitative_result} presenting qualitative comparison to our competitors.


We also evaluate the proposed method on the online KITTI benchmark server with a model trained using KITTI's official split.
Apart from the training set, all other settings remain the same as in the experiment using KITTI's Eigen split.
We trained Ours-DenseNet for 50 epochs with 28,654 image-ground truth pairs sampled from the official training and validation set.
As shown in Table \ref{tb:kitti_online_benchmark}, our method outperforms all the published works.

\begin{table*}[t]
	\centering
	\resizebox{2\columnwidth}{!}{%
		\begin{tabular}{r|r|ccc|ccccc}
			\hline
			\multicolumn{1}{c|}{\multirow{2}{*}{Variant}} & \multicolumn{1}{c|}{\multirow{2}{*}{\# Params}} & \multicolumn{3}{c|}{\textit{higher is better}} & \multicolumn{5}{c}{\textit{lower is better}} \\ \cline{3-10} 
			\multicolumn{1}{c|}{} &  \multicolumn{1}{c|}{} & $\delta<1.25$ & $\delta<1.25^2$ & $\delta<1.25^3$ & Abs Rel & Sq Rel & RMSE & RMSE \textit{log} & log10 \\ \hline
			MobileNetV2 \cite{sandler2018mobilenetv2} & 16.3M  & 0.860 & 0.974 & 0.993 & 0.121 & 0.080 & 0.431 & 0.156 & 0.052 \\
			ResNet-50 \cite{he2016deep}               & 49.5M  & 0.865 & 0.975 & 0.993 & 0.119 & 0.075 & 0.419 & 0.152 & 0.051 \\
			ResNet-101 \cite{he2016deep}              & 68.5M  & 0.871 & 0.977 & \bf{0.995} & 0.113 & 0.068 & 0.407 & 0.148 & 0.049 \\
			ResNext-50 \cite{xie2017cvpr}             & 49.0M  & 0.867 & 0.977 & \bf{0.995} & 0.116 & 0.070 & 0.414 & 0.150 & 0.050 \\
			ResNext-101 \cite{xie2017cvpr}            & 112.8M & 0.880 & 0.977 & 0.994 & 0.111 & 0.069 & 0.399 & 0.145 & 0.048 \\
			DenseNet-121 \cite{huang2017densely}      & 21.2M  & 0.871 & 0.977 & 0.993 & 0.118 & 0.072 & 0.410 & 0.149 & 0.050 \\
			DenseNet-161 \cite{huang2017densely}      & 47.0M  & \bf{0.885} & \bf{0.978} & 0.994 & \bf{0.110} & \bf{0.066} & \bf{0.392} & \bf{0.142} & \bf{0.047} \\ \hline
		\end{tabular}
	}
	\caption{Experimental results using NYU Depth V2 with various base netowrks.}
	\label{tb:various_backbones_nyu}
\end{table*}

\begin{table*}[t]
	\centering
	\resizebox{2\columnwidth}{!}{%
		\begin{tabular}{r|r|ccc|ccccc}
			\hline
			\multicolumn{1}{c|}{\multirow{2}{*}{Variant}} & \multicolumn{1}{c|}{\multirow{2}{*}{\# Params}} & \multicolumn{3}{c|}{\textit{higher is better}} & \multicolumn{5}{c}{\textit{lower is better}} \\ \cline{3-10} 
			\multicolumn{1}{c|}{} &  \multicolumn{1}{c|}{} & $\delta<1.25$ & $\delta<1.25^2$ & $\delta<1.25^3$ & Abs Rel & Sq Rel & RMSE & RMSE \textit{log} & log10 \\ \hline
			ResNet-50 \cite{he2016deep}               & 49.5M  & 0.954 & 0.992 & \bf{0.998} & 0.061 & 0.250 & 2.803 & 0.098 & 0.027 \\
			ResNet-101 \cite{he2016deep}              & 68.5M  & 0.954 & \bf{0.993} & \bf{0.998} & 0.061 & 0.261 & 2.834 & 0.099 & 0.027 \\			
			DenseNet-121 \cite{huang2017densely}      & 21.2M  & 0.951 & \bf{0.993} & \bf{0.998} & 0.063 & 0.256 & 2.850 & 0.100 & 0.028 \\
			DenseNet-161 \cite{huang2017densely}      & 47.0M  & 0.955 & \bf{0.993} & \bf{0.998} & 0.060 & 0.249 & 2.798 & 0.096 & 0.027 \\
			ResNext-50 \cite{xie2017cvpr}             & 49.0M  & 0.954 & \bf{0.993} & \bf{0.998} & 0.061 & 0.245 & 2.774 & 0.098 & 0.027 \\
			ResNext-101 \cite{xie2017cvpr}            & 112.8M & \bf{0.956} & \bf{0.993} & \bf{0.998} & \bf{0.059} & \bf{0.241} & \bf{2.756} & \bf{0.096} & \bf{0.026} \\ \hline
		\end{tabular}
	}
	\caption{Experimental results using KITTI's Eigen split with various base netowrks. In this experiment, we set the capturing range to $0-80m$.}
	\label{tb:various_backbones_kitti}	
\end{table*}

\subsection{Ablation Study}
\label{sec:ablation_study}
Here, we conduct evaluations with variants of our network to see the effectiveness of the core factors.
From the baseline network, which only consists of the base network (\textit{i.e., ResNet-101}) and a simple upsampling layer, we increment the network with the core modules to see how the added factor improves the performance.
The result is given in Table \ref{tb:ablation_study}.
As the core factors are added, the overall performance is improved, and the most significant improvement is made by adding the proposed local planar guidance layers.
Please note that the LPG layers only require additional 0.1M trainable parameters used by \textit{1x1 reduction} layers.
The final improvement comes from using the training loss defined in Equation \ref{eq:loss}.
Benefited from the robust base network \textit{DenseNet-161}, ours achieves state-of-the-art performance with the significant margin while it requires the less number of parameters than the baseline.

\subsection{Experiments with Various Base Networks}
\label{sec:experiments_with_various_base_networks}
Because the proposed network adopts existing models as an encoder for dense feature extraction, it is desirable to see how the performance varies with various base networks that are widely used for similar applications.
By changing the encoder with various models while other settings remained, we experimented with the proposed method using both of the NYU Depth V2 and KITTI's Eigen split, and provide the result in Tables \ref{tb:various_backbones_nyu} and \ref{tb:various_backbones_kitti}.
Note that ResNet-101, ResNext-101 and DenseNet-161 are identical to Ours-ResNet, Ours-ResNext and Ours-DenseNet, respectively, in Tables \ref{tb:nyu} and \ref{tb:result_kitti_eigen}.
Interestingly, for the NYU Depth V2 dataset, DenseNet-161 results in the best performance, while for KITTI's Eigen split, ResNext-101 achieves the state-of-the-art result.
We consider this as an effect of the relatively lower variance of the data distribution in the indoor scenes of the NYU Depth V2 dataset, which can lead to a degeneration of performance with very deep models such as ResNext-101 in this experiment.
Also, it is notable that our MobileNetV2-based model results in performance drop about only 3\% for inlier measures and less than 15\% drop for accuracy measures while it contains less than half the number of parameters and shows about three times speedup when compared to our model based on the DenseNet-161.

\subsection{Qualitative Result}
\label{sec:qualitative_result}
\begin{figure*}
	\centering
	\begin{subfigure}{.18\linewidth}
		\centering
		\includegraphics[width=\linewidth, height=2.1cm]{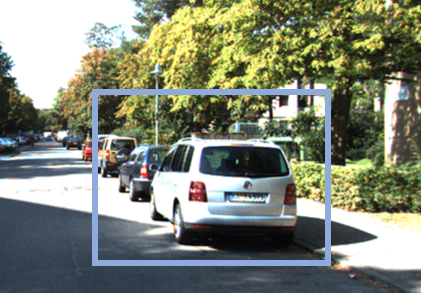}
	\end{subfigure}
	\hspace{1em}
	\begin{subfigure}{.18\linewidth}
		\centering
		\includegraphics[width=\linewidth, height=2.1cm]{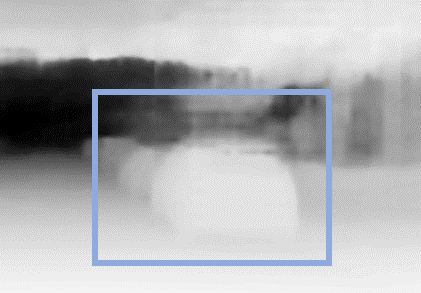}
	\end{subfigure}
	\hspace{1em}
	\begin{subfigure}{.18\linewidth}
		\centering
		\includegraphics[width=\linewidth, height=2.1cm]{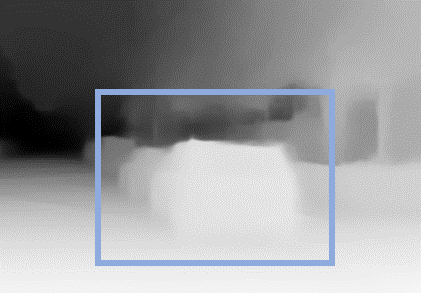}
	\end{subfigure}
	\hspace{1em}
	\begin{subfigure}{.18\linewidth}
		\centering
		\includegraphics[width=\linewidth, height=2.1cm]{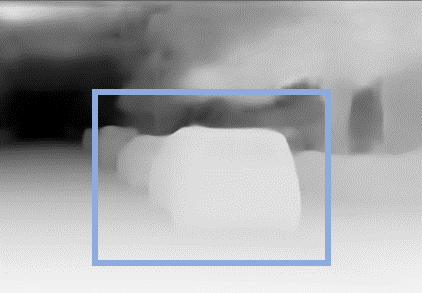}
	\end{subfigure}
	
	\begin{subfigure}{.18\linewidth}
		\centering
		\includegraphics[width=\linewidth, height=2.1cm]{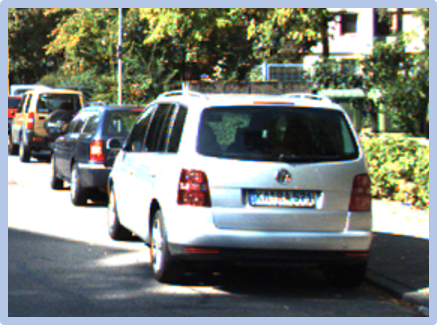}
	\end{subfigure}
	\hspace{1em}
	\begin{subfigure}{.18\linewidth}
		\centering
		\includegraphics[width=\linewidth, height=2.1cm]{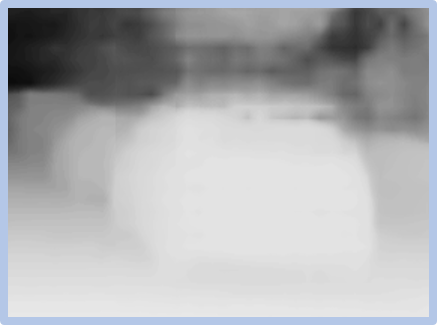}
	\end{subfigure}
	\hspace{1em}
	\begin{subfigure}{.18\linewidth}
		\centering
		\includegraphics[width=\linewidth, height=2.1cm]{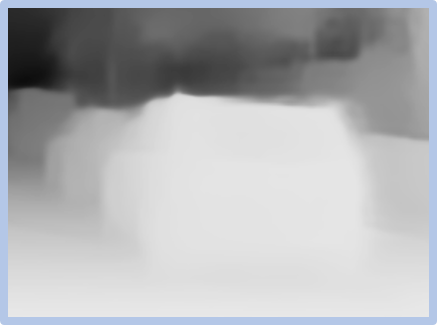}
	\end{subfigure}
	\hspace{1em}
	\begin{subfigure}{.18\linewidth}
		\centering
		\includegraphics[width=\linewidth, height=2.1cm]{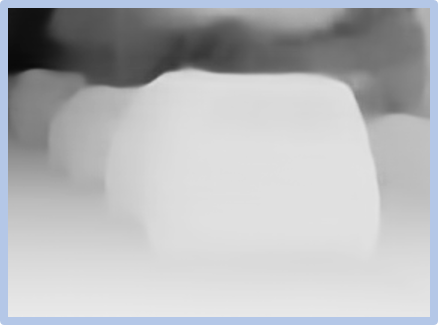}
	\end{subfigure}
	\vspace{0.15cm}
	
	\begin{subfigure}{.18\linewidth}
		\centering
		\includegraphics[width=\linewidth, height=2.1cm]{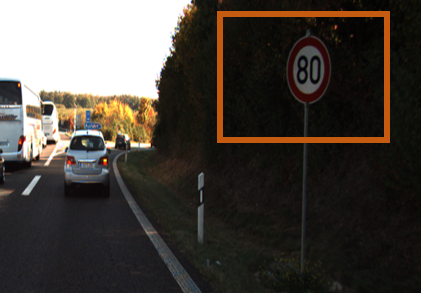}
	\end{subfigure}
	\hspace{1em}
	\begin{subfigure}{.18\linewidth}
		\centering
		\includegraphics[width=\linewidth, height=2.1cm]{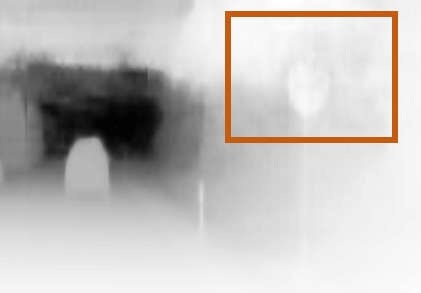}
	\end{subfigure}
	\hspace{1em}
	\begin{subfigure}{.18\linewidth}
		\centering
		\includegraphics[width=\linewidth, height=2.1cm]{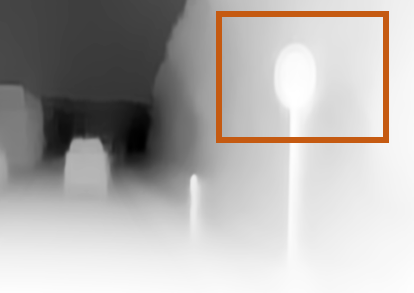}
	\end{subfigure}
	\hspace{1em}
	\begin{subfigure}{.18\linewidth}
		\centering
		\includegraphics[width=\linewidth, height=2.1cm]{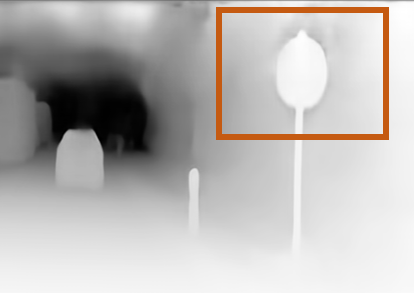}
	\end{subfigure}
	
	\begin{subfigure}{.18\linewidth}
		\centering
		\includegraphics[width=\linewidth, height=2.1cm]{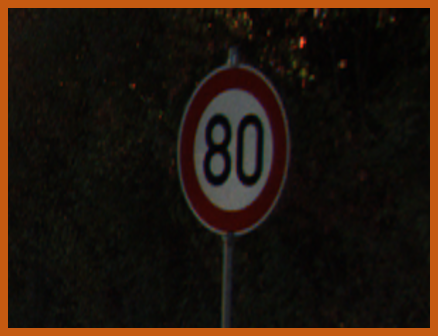}
		\caption{Input}
	\end{subfigure}
	\hspace{1em}
	\begin{subfigure}{.18\linewidth}
		\centering
		\includegraphics[width=\linewidth, height=2.1cm]{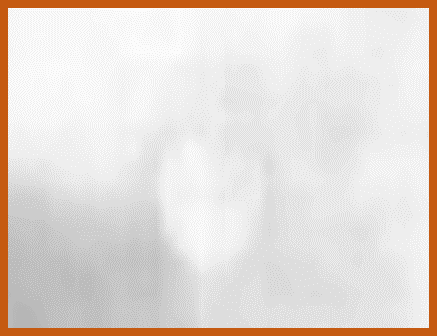}
		\caption{Fu et al. \cite{fu2018deep}}		
	\end{subfigure}
	\hspace{1em}
	\begin{subfigure}{.18\linewidth}
		\centering
		\includegraphics[width=\linewidth, height=2.1cm]{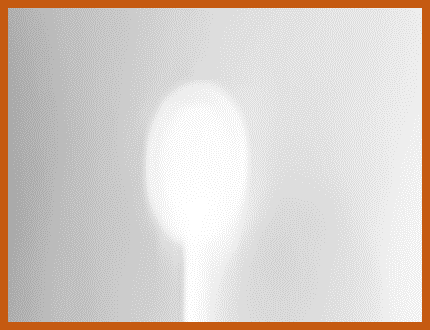}
		\caption{Yin et al. \cite{yin2019enforcing}}
	\end{subfigure}
	\hspace{1em}
	\begin{subfigure}{.18\linewidth}
		\centering
		\includegraphics[width=\linewidth, height=2.1cm]{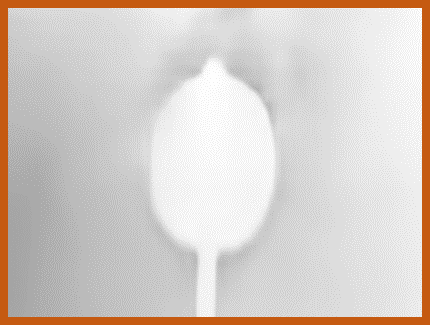}
		\caption{Ours}
	\end{subfigure}
	\caption{\textbf{Qualitative results on the KITTI Eigen test split.} The proposed method results clearer boundaries from the vehicle and traffic sign.}
	\label{fg:kitti_qualitative_result}
\end{figure*}

\begin{figure*}
	\centering
	\begin{subfigure}{.18\linewidth}
		\centering
		\includegraphics[width=\linewidth, height=2.1cm]{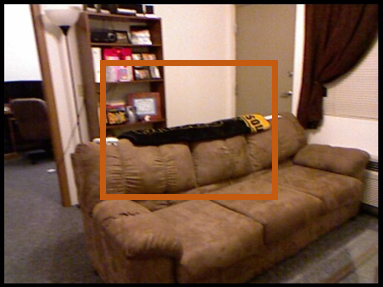}
	\end{subfigure}
	\hspace{1em}
	\begin{subfigure}{.18\linewidth}
		\centering
		\includegraphics[width=\linewidth, height=2.1cm]{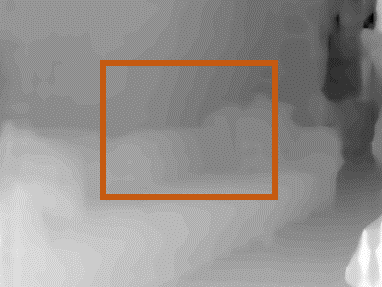}
	\end{subfigure}
	\hspace{1em}
	\begin{subfigure}{.18\linewidth}
		\centering
		\includegraphics[width=\linewidth, height=2.1cm]{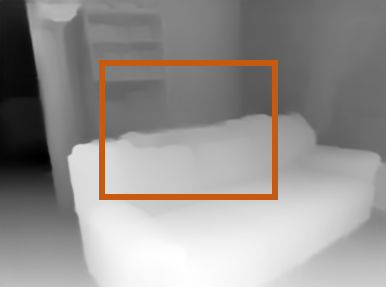}
	\end{subfigure}
	\hspace{1em}
	\begin{subfigure}{.18\linewidth}
		\centering
		\includegraphics[width=\linewidth, height=2.1cm]{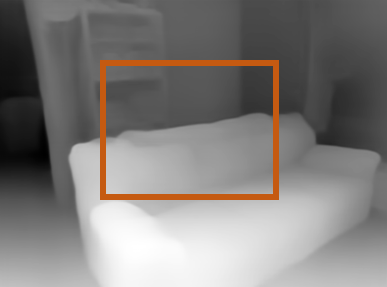}
	\end{subfigure}
	
	\begin{subfigure}{.18\linewidth}
		\centering
		\includegraphics[width=\linewidth, height=2.1cm]{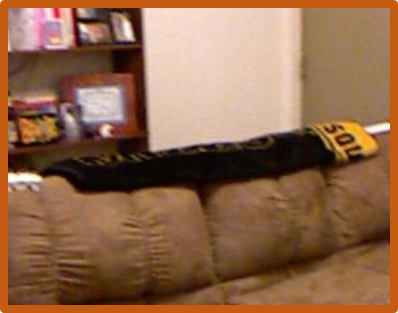}
	\end{subfigure}
	\hspace{1em}
	\begin{subfigure}{.18\linewidth}
		\centering
		\includegraphics[width=\linewidth, height=2.1cm]{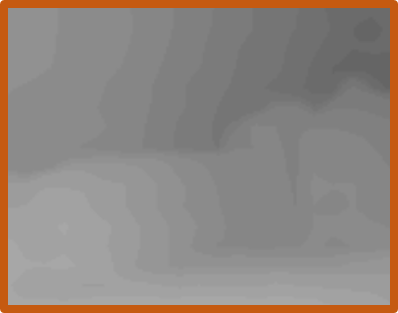}
	\end{subfigure}
	\hspace{1em}
	\begin{subfigure}{.18\linewidth}
		\centering
		\includegraphics[width=\linewidth, height=2.1cm]{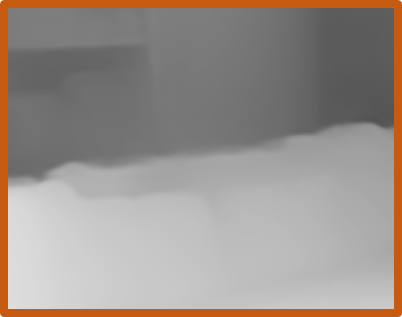}
	\end{subfigure}
	\hspace{1em}
	\begin{subfigure}{.18\linewidth}
		\centering
		\includegraphics[width=\linewidth, height=2.1cm]{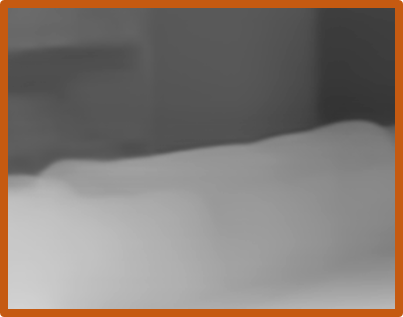}
	\end{subfigure}
	\vspace{0.15cm}
	
	\begin{subfigure}{.18\linewidth}
		\centering
		\includegraphics[width=\linewidth, height=2.1cm]{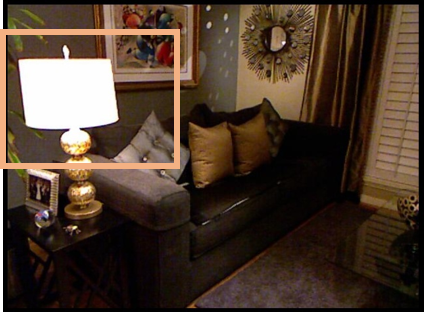}
	\end{subfigure}
	\hspace{1em}
	\begin{subfigure}{.18\linewidth}
		\centering
		\includegraphics[width=\linewidth, height=2.1cm]{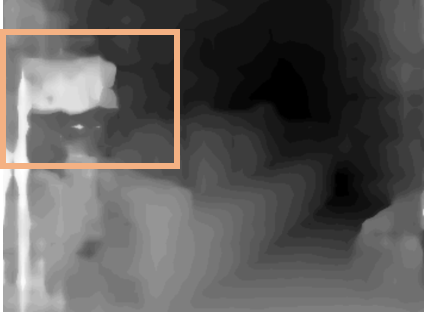}
	\end{subfigure}
	\hspace{1em}
	\begin{subfigure}{.18\linewidth}
		\centering
		\includegraphics[width=\linewidth, height=2.1cm]{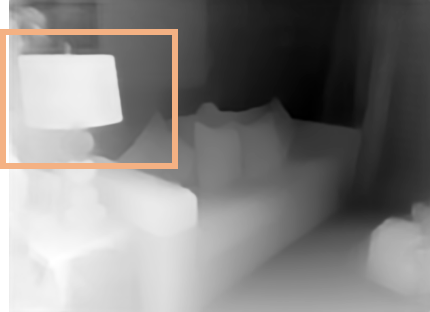}
	\end{subfigure}
	\hspace{1em}
	\begin{subfigure}{.18\linewidth}
		\centering
		\includegraphics[width=\linewidth, height=2.1cm]{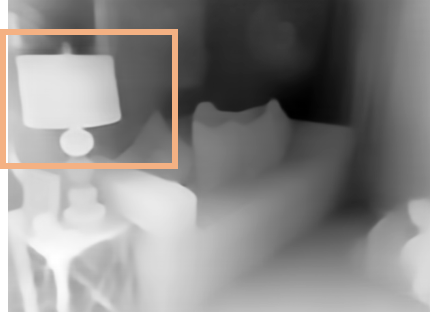}
	\end{subfigure}
	
	\begin{subfigure}{.18\linewidth}
		\centering
		\includegraphics[width=\linewidth, height=2.1cm]{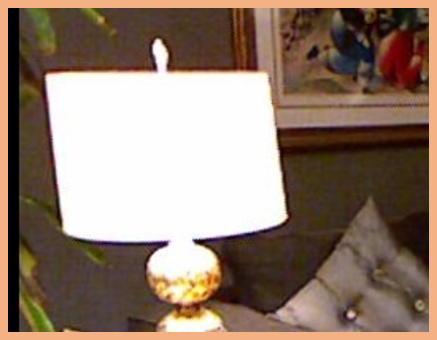}
		\caption{Input}
	\end{subfigure}
	\hspace{1em}
	\begin{subfigure}{.18\linewidth}
		\centering
		\includegraphics[width=\linewidth, height=2.1cm]{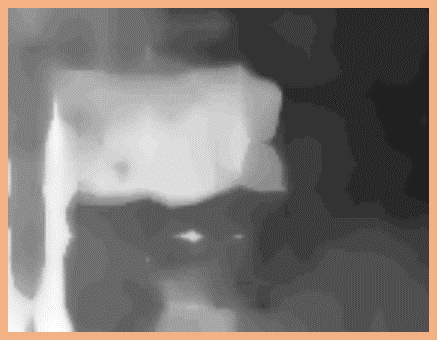}
		\caption{Fu et al. \cite{fu2018deep}}
	\end{subfigure}
	\hspace{1em}
	\begin{subfigure}{.18\linewidth}
		\centering
		\includegraphics[width=\linewidth, height=2.1cm]{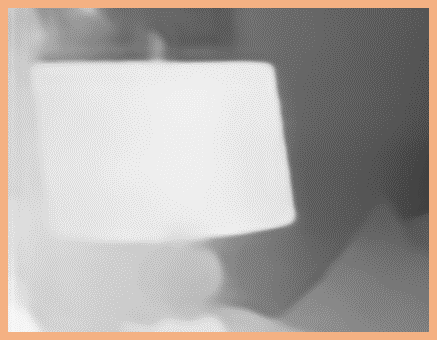}
		\caption{Yin et al. \cite{yin2019enforcing}}
	\end{subfigure}
	\hspace{1em}
	\begin{subfigure}{.18\linewidth}
		\centering
		\includegraphics[width=\linewidth, height=2.1cm]{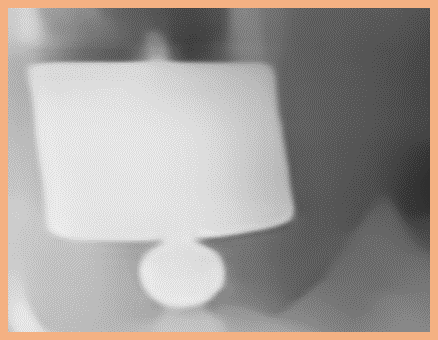}
		\caption{Ours}
	\end{subfigure}
	\caption{\textbf{Qualitative results on the NYU Depth V2 test split.} While the method from Yin et al \cite{yin2019enforcing} show competitive results to ours, our method achieves more distinctive results especially on object boundaries. }
	\label{fg:nyu_qualitative_result}
\end{figure*}

Finally, we discuss qualitative results from ours and competing works.
As we can see from Figures \ref{fg:kitti_qualitative_result} and \ref{fg:nyu_qualitative_result}, ours show much more precise object boundaries.
However, in results from experiments using KITTI, we can see artifacts in the sky or upper part of the scenes.
We consider this as a consequence of the very sparse ground truth depth data.
Because certain regions are lacking valid depth values across the dataset, the network cannot be appropriately trained for those regions.

\section{Conclusion}
\label{sec:conclusion}
In this work, we have presented a supervised monocular depth estimation network and achieved state-of-the-art results.
Benefiting from recent advances in deep learning, we design a network architecture that uses novel local planar guidance layers, giving an explicit relation from internal feature maps to the desired prediction for better training of the network.
By deploying the proposed layers on multiple stages in the decoding phase, we have gained a significant improvement and shown several experimental results on challenging benchmarks to verify it.
However, in experiments with the KITTI dataset, we observe frequent artifacts on the upper part of the scenes.
We analyze this as an effect of the high sparsity of the ground truth across the dataset.
As a consequence, we plan to investigate adopting into our framework a photometric reconstruction loss, which can provide far denser supervision to improve the performance further.

{\small
	\bibliographystyle{ieee}
	\bibliography{bts_main}
}

\end{document}